\documentclass[conference]{IEEEtran}
\IEEEoverridecommandlockouts
 \usepackage[table,xcdraw]{xcolor}
\usepackage{amsmath}
\usepackage[colorlinks]{hyperref}
\usepackage{amsfonts}
\usepackage{graphicx}
\usepackage{textcomp}
\usepackage{xcolor}
\usepackage{multirow}
\usepackage{soul}
\usepackage{tikz}
\usepackage{algorithm}
\usepackage{algpseudocode}
\usepackage{colortbl}
\usepackage{bm}
\usepackage{url}
\usepackage{float}
\usepackage{hyperref}
\usepackage{filecontents}
\usepackage{todonotes}
\usepackage[table]{xcolor}
\usepackage{booktabs}
\usepackage{caption}
\usepackage{tablefootnote}
\usepackage{threeparttable}

\definecolor{hBase}{HTML}{E8E8E8}   
\definecolor{hV3}{HTML}{FFF9C4}     
\definecolor{hV4}{HTML}{B3E5FC}     
\definecolor{hV5}{HTML}{C8E6C9}     
\definecolor{cZero}{HTML}{FFCDD2}   
\definecolor{cFull}{HTML}{C8E6C9}   
\definecolor{cPart}{HTML}{FFF9C4}   
\definecolor{tC1}{HTML}{B2DFDB}
\definecolor{tC2}{HTML}{FFE0B2}
\definecolor{tC3}{HTML}{E1BEE7}
 
\newcommand{\Szero}{\cellcolor{cZero}0.0}
\newcommand{\Sfull}{\cellcolor{cFull}\textbf{100.0}}
\newcommand{\Spart}[1]{\cellcolor{cPart}#1}
\newcommand{\TC}[1]{%
  \ifnum#1=1\cellcolor{tC1}\textbf{C1}\fi%
  \ifnum#1=2\cellcolor{tC2}\textbf{C2}\fi%
  \ifnum#1=3\cellcolor{tC3}\textbf{C3}\fi%
}

\newcommand{\cmmnt}[1]{}  

\newcommand\encircle[1]{%
\tikz[baseline=(X.base)] 
  \node (X) [draw, scale=0.75, shape=circle, inner sep=0, fill=black, text=white, minimum size=0em] {\strut #1};}

\usepackage{pifont}
\makeatletter 
\newcommand\semiHuge{\@setfontsize\semiHuge{22.72}{27.38}}
\makeatother

\usepackage[a4paper, total={184mm,239mm}]{geometry}
\def\BibTeX{{\rm B\kern-.05em{\sc i\kern-.025em b}\kern-.08em
    T\kern-.1667em\lower.7ex\hbox{E}\kern-.125emX}}

\begin{document}

\title{\semiHuge PICasso: An AI-Enabled Design Framework for Autonomous Optimization of Silicon Photonic Devices\vspace{-0.5em}}

\author{Deepak Vungarala$^{1}$,
Deniz Najafi$^{1}$,
Abdulrahman Aljoudi$^{2}$, 
Zahra Ghanaatian$^{3}$,
Navid Khoshavi$^{4}$,\\
Gourav Datta$^{5}$,
Arman Roohi$^{6}$,
Mahdi Nikdast$^{3}$,
Shaahin Angizi$^{1}$\\
\small
$^1$New Jersey Institute of Technology, USA
$^2$University of Michigan, USA
$^3$Colorado State University, USA
$^4$AMD, USA\\
$^5$Case Western Reserve, USA
$^6$University of Illinois, Chicago, USA
\\
E-mails: \{dv336,shaahin.angizi\}@njit.edu\\
\vspace{-3em}}

\maketitle

\begin{abstract}
We present \textsc{PICasso}, an AI-assisted framework for automated synthesis, verification, and optimization of photonic integrated circuits (PICs) from natural-language specifications.
PICasso couples a structured NL$\rightarrow$YAML$\rightarrow$GDS generation pipeline with PDK aware knowledge injection, automated placement and routing, DRC/LVS validation, and SAX-based photonic simulation. To systematically evaluate AI-driven photonic design, we introduce \textsc{PIC-Set}, a benchmark of 36 parameterized PIC design tasks spanning core photonic primitives and multi-component circuits. Using PIC-Set, we benchmark several state-of-the-art Large Language Models (LLMs) under a unified evaluation protocol, including new metrics such as structural and functional $\mathrm{Spec}@k$, optimization efficiency, and robustness under perturbations. Across the benchmark, PICasso significantly improves end-to-end specification satisfaction compared to vanilla LLM generation. Structural $\mathrm{Spec}@3$ reaches up to 92.7\% and functional $\mathrm{Spec}@3$ up to 52\% on high-complexity circuits. In addition, PICasso consistently reduces circuit insertion loss, lowering the mean loss from 4.98\,dB to 3.25\,dB (1.74\,dB improvement) through simulation-guided optimization. These results demonstrate that structured domain constraints, physical verification, and simulation feedback transform LLMs from brittle netlist generators into practical PIC design agents capable of producing manufacturable layouts with competitive runtimes relative to manual GUI-based workflows.
 \textit{Our code and dataset are released at} \url{https://anonymous.4open.science/r/PICasso-7D52}
\end{abstract}

\maketitle
\pagestyle{plain}

 \vspace{-0.75em} 
\section{Introduction}

Silicon photonic integrated circuits (PICs) underpin emerging applications in optical interconnects, LiDAR, and quantum photonics, with the addressable market projected to exceed \$7 billion within five years~\cite{marketsandmarkets2025}. Yet the design toolchain has not kept pace: circuit complexity is scaling faster than the manual, GUI-centric workflows that dominate current photonic design automation (PDA). However, today's PIC design workflows are fundamentally misaligned with the scale and rigor demanded by this rapid market growth. Current photonic design automation (PDA) tools remain predominantly GUI-centric \cite{layoutcopilot}, which makes them ill-suited for the systematic development required when circuits scale to thousands of elements. Engineers manually draw and wire every component, a process that lacks version control, reproducibility, and the parametric flexibility essential for exploring design space \cite{interactive1} \cite{interactive2}. As circuit complexity increases, this manual approach becomes untenable: placement and routing effort scales super-linearly, amplifying human error; parametrically defined building blocks for optimization are complex or impossible to express in graphical interfaces; and critically, the design, simulation, and verification stages remain fragmented across incompatible commercial tools (Lumerical \cite{ansyslumerical}, Ansys \cite{ansys}) and open-source alternatives (Scattering Analysis eXtension-SAX \cite{SAX}, gdsfactory \cite{gdsfactory}), forcing repeated manual translation that introduces errors and delays time-to-market. 

Large Language Models (LLMs) are increasingly being applied to hardware design, enabling the generation, optimization, and verification of Hardware Description Language (HDL) / High-Level Synthesis (HLS) code. Works such as VeriGen \cite{thakur2023verigen}, ChatEDA \cite{he2023chateda}, and Autochip \cite{thakur2023autochip} integrate LLMs into RTL-to-GDSII flows, automate assertion creation, and leverage simulation feedback for precise code synthesis. AnalogCoder \cite{lai2024analogcoder} and SPICEPilot~\cite{vungarala2024spicepilot} pioneer the generation of analog circuits via in-context learning. In PIC design, research is sparse: early work generated FDTD scripts for PCSEL simulation~\cite{li:hal-04175312}, translated natural language into GDSII layouts~\cite{liu2024towards}. PIC-Bench \cite{wu2025picbench} has recently established the first open-source benchmark for PIC design, offering 24 canonical circuit design tasks spanning filters, switches, and interconnect devices. 

While such benchmarking efforts represent an important step toward standardizing PIC evaluation, current LLM-based approaches remain fundamentally limited. Most rely on heuristic prompt engineering, ad hoc trial-and-error workflows, and primitive error-repair loops that fail to leverage the reasoning capabilities of modern foundation models. Moreover, existing systems incorporate only minimal domain knowledge, resulting in designs that are often syntactically valid yet physically unrealizable or functionally incorrect.
These limitations motivate three central research questions for the next generation of PDA \textit{\textbf{RQ1}}:~To what extent can 
LLMs generate functionally correct PIC designs when guided by explicit photonic domain knowledge and structured constraints? \textit{\textbf{RQ2}}:~Can device-level optimization, traditionally a highly manual process, be automated end-to-end through LLM-driven generation and simulation-guided refinement? \textit{\textbf{RQ3:}} How can device- and circuit-level optimization be  unified in a single closed loop to enable research-grade PIC  design with minimal human intervention?
To address these questions, we introduce the following contributions in this manuscript.

(i) We propose \textsc{PICasso}, a fully automated LLM-assisted PIC design framework integrating structured YAML generation, PDK-aware knowledge injection, pilot validation, DRC/LVS  verification, and closed-loop regeneration. $(ii)$ We design \textbf{PIC-Set}, a curated, open-source dataset comprising over 36 parameterized component- and circuit-level design tasks. (iii) We implement an automated two-stage optimization loop that  unifies device-level geometry tuning with circuit-level interference  control via SAX simulations, representing the first end-to-end LLM-driven optimization flow for PICs. (iv) We propose evaluation metrics: structural and functional $\mathrm{Spec}@k$, optimization efficiency, and use them to benchmark several state-of-the-art LLMs across the full PIC-Set.

Together, PIC-Set and PICasso establish a first-of-its-kind foundation for evaluating and advancing LLM-driven PIC design. They demonstrate that, when paired with structured domain constraints and physical verification, modern LLMs can transition from producing sporadically valid designs to reliably generating manufacturable, optimized PICs at scale.


\section{Background}
\noindent\textbf{LLM for Hardware Design.} Recent studies demonstrate the potential of LLMs for generating HDL and HLS designs. Frameworks such as VeriGen \cite{thakur2023verigen} and ChatEDA \cite{he2023chateda} automate stages of the RTL-to-GDSII workflow, while AssertLLM \cite{fang2024assertllm} and UVLLM \cite{hu2024uvllm} leverage specialized LLM pipelines to generate SystemVerilog assertions and automate RTL verification and repair. LLM-driven hardware generation and optimization have also been explored in ChipGPT \cite{chang2023chipgpt} and AutoChip \cite{thakur2023autochip}, where simulation feedback iteratively refines Verilog designs. Several efforts focus on datasets and design automation frameworks. TPU-Gen \cite{11106005} and MG-Verilog \cite{zhang2024mgverilog} introduce large-scale datasets for automated hardware generation, while Chip-Chat \cite{blocklove2023chip}, MEV-LLM \cite{nadimi2024multi}, and DeepCircuitX \cite{li2025deepcircuitx} explore interactive, multi-expert, and Chain-of-Thought–enhanced approaches for RTL generation and early PPA estimation. Other systems, including RTLLM \cite{lu2023rtllm}, GPT4AIGChip \cite{10323953}, and LLMCompass \cite{zhang2024llmcompass}, further demonstrate improved design productivity and architectural exploration. Despite these advances, challenges such as prompt optimization, domain-specific datasets, model fine-tuning, and hallucinations continue to limit LLM-based EDA pipelines \cite{he2023chateda}. Consequently, many works rely on in-context learning (ICL) to improve performance \cite{dai2022can}. In the analog domain, AnalogCoder \cite{lai2024analogcoder}, SPICEPilot \cite{vungarala2024spicepilot}, and Masala-CHAI \cite{bhandari2025masalachailargescalespicenetlist} represent early attempts at circuit generation through prompt engineering and ICL, while emerging directions explore LLM-assisted hardware paradigms such as in-memory computing with LIMCA \cite{vungarala2025limca}.

\noindent\textbf{Automating PIC Design.}
A substantial body of work has focused on inverse design and related semi-automated or AI-based methods for PIC design \cite{maclellan2024inverse, li2023deep,khan2019photonic}. For example, topology- and adjoint-based inverse-design methods have been shown to explore high-dimensional device parameter spaces efficiently and yield markedly more compact, higher-performance photonic devices than conventional forward sweeps \cite{shen2024recent}. In addition, deep-learning and generative-model approaches such as convolutional neural networks, autoencoders, and reinforcement-learning frameworks have recently been applied to the inverse design of silicon photonic devices, enabling rapid mapping from target optical functionality to geometry \cite{baek2025recent}. Further advancing the automation frontier, coupled device-level and layout-level frameworks integrate fabrication-aware inverse-design engines with automated placement and routing (PnR) tools, thereby lowering human intervention and improving scalability of PIC workflows \cite{zhou2025toward}. Collectively, these semi-automated pipelines demonstrate that AI-driven, simulation-based loops can reduce manual expert tuning and enable broader exploration of design space, thus promising more autonomous PIC design systems. 

Research on applying LLMs to PIC design remains scarce, with only a few preliminary studies addressing this area. The use of LLMs to generate FDTD scripts for simulating a photonic crystal surface-emitting laser (PCSEL) structure, followed by AI-based optimization of the device parameters, has been shown in \cite{li:hal-04175312}. However, their workflow was not fully autonomous, relying heavily on expert intervention to define specifications and correct coding errors iteratively. Liu et al. \cite{liu2024towards} introduced an automated pipeline capable of translating natural-language descriptions into Python code for generating GDSII layouts via an open-source library. While promising, their evaluation was limited to seven relatively simple devices, leaving questions about scalability and generalization to more complex designs. The most promising work, i.e., PIC-Bench \cite{wu2025picbench}, introduced an open-source benchmark for PIC design, comprising 24 distinct photonic integrated circuit design problems. The framework was assessed using state-of-the-art commercial LLMs and demonstrated relatively acceptable syntax correctness rates; however, the functionality outcomes were highly unsatisfactory. With certain restrictions applied, Gemini~1.5~Pro’s functionality score improved from 8.33\% to 21.67\% in Pass@1, and from 12.50\% to 37.50\% in Pass@5 at best. 

\begin{figure}[t]
    \centering
    \includegraphics[width=1.01\linewidth]{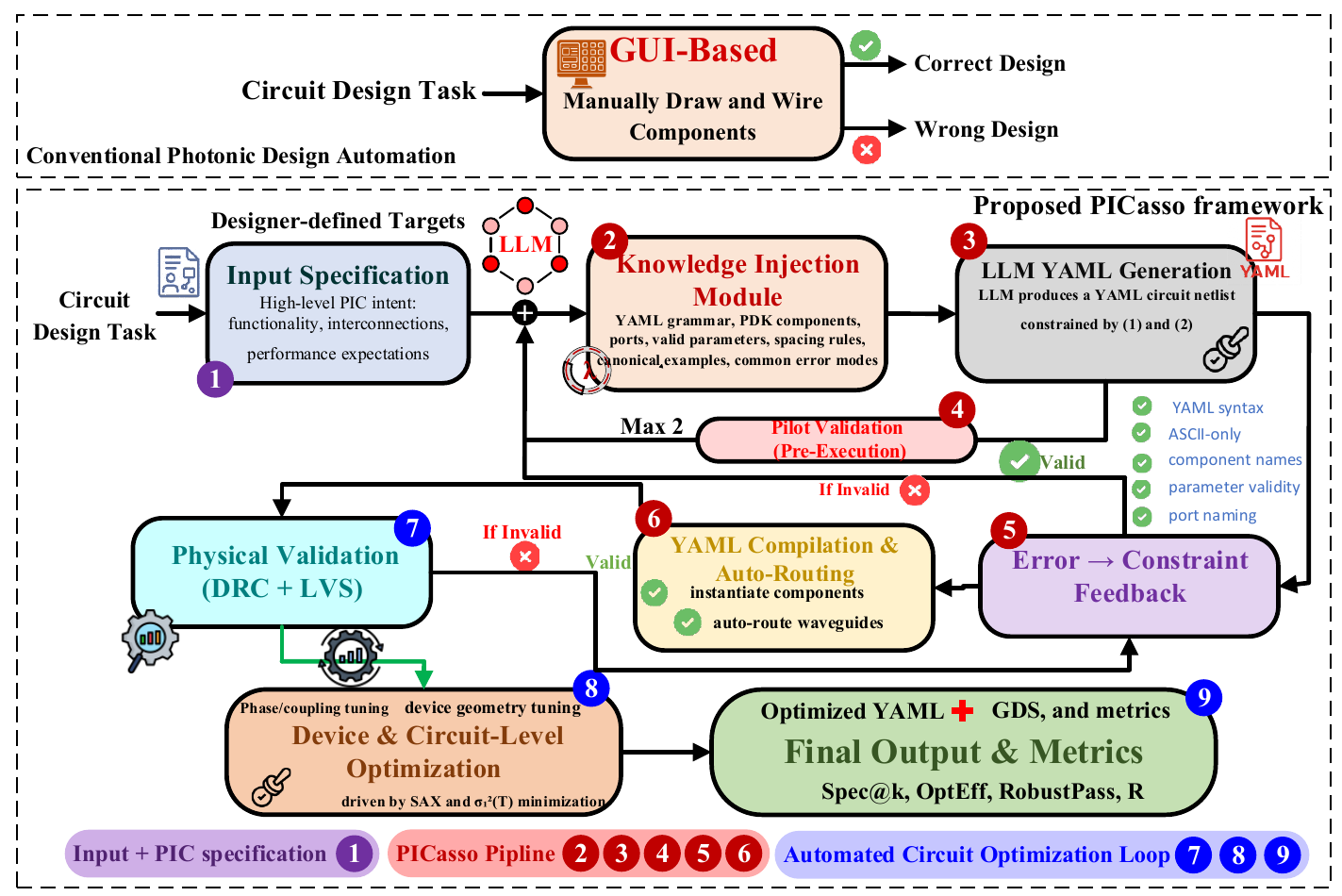} \vspace{-2em}
    \caption{Picasso Framework.}
    \label{fig:framework} \vspace{-2em}
\end{figure}

\section{PICasso } \label{sec:framework} 
The \textsc{PICasso} framework (Fig.~\ref{fig:framework}) provides an automated, constraint-driven synthesis pipeline that maps high-level photonic intent to manufacturable, PDK-compliant layouts. Its design is anchored in two core principles: (i) \emph{prompt-constrained generation}, where the LLM receives a structured, domain-informed specification that restricts its output space to valid photonic circuits, and (ii) \emph{feedback-driven refinement}, where syntactic, semantic, and physical violations are automatically identified and translated into corrective constraints for iterative regeneration. Together, these mechanisms bridge the gap between natural-language circuit descriptions and physically verifiable photonic implementations. A summary of quantitative evaluation metrics used throughout the pipeline is provided in Sec.~\ref{sec:metrics}.

\subsection{Framework}
 From Fig.~\ref{fig:framework} the synthesis process begins with a user-defined description of circuit functionality, interconnections, and optional performance goals such as insertion loss or splitting ratio \encircle{1}. This specification is augmented by a \emph{knowledge injection module} \encircle{2}, which supplies the LLM with formal YAML grammar rules, valid component and port definitions from the target PDK, admissible parameter sets obtained via \texttt{inspect.signature}, spacing requirements, and a curated library of canonical design examples. The injection also enumerates common failure modes—such as illegal port names, invalid MMI parameters, and insufficient spacing—thereby constraining the LLM to generate foundry-compatible netlists rather than free-form layouts.

A dedicated \emph{pilot prompt} \encircle{3} performs pre-execution validation on each generated YAML file. This validator enforces strict ASCII-only formatting, YAML schema compliance, correct parameter usage, port name consistency, adherence to minimum spacing rules, and the completeness of placement and routing directives. Any violations encountered at this stage are extracted, distilled into concise diagnostic messages, and appended to the pilot prompt \encircle{4}, forming an adaptive repair loop that persists across the local synthesis run while maintaining readability and prompt stability.

Once the YAML netlist satisfies pilot validation, it is compiled using \texttt{gf.read.from\_yaml()} \encircle{5}, which instantiates all components, performs geometric placement, and invokes gdsfactory’s auto-routing engine. The router generates fabrication-feasible waveguide paths that respect bend-radius limits, cross-section constraints, and PDK-defined spacing rules, leveraging systematic handling of Euler bends, tapers, straight segments, and Manhattan routing primitives. The resulting layout is then subjected to two forms of physical verification. A \emph{design rule check} (DRC) \encircle{6} evaluates width, spacing, enclosure, and curvature constraints using the \texttt{generic\_tech} PDK and KLayout's rule engine. When computationally tractable, a \emph{layout-versus-schematic} (LVS) comparison verifies connectivity consistency between the layout-extracted netlist and the original YAML, preventing missing or shorted waveguides. Any DRC or LVS violations generate structured feedback that is returned to the pilot mechanism \encircle{7}, guiding targeted corrections for subsequent LLM generations.

Designs that pass physical validation enter the \emph{optimization stage} \encircle{8}, where device-level geometry tuning and circuit-level phase and coupling refinement are performed. PICasso employs SAX-based simulations to evaluate optical performance, enabling optimization toward lower insertion loss, improved balance, and enhanced functional fidelity. The finalized outputs include an optimized YAML schematic, the corresponding GDS layout, and performance metadata used in Sec.~\ref{sec:metrics}. This closed-loop synthesis–validation–optimization pipeline enables rapid, minimally supervised generation of manufacturable photonic circuits, establishing a scalable foundation for next-generation photonic EDA workflows.

As summarized in Table~\ref{tab:comparison}, prior LLM-driven PIC design frameworks remain constrained in several critical ways. PICBench operates exclusively at the netlist level and provides no physical verification or layout generation, while PhIDO introduces NL-to-layout generation but lacks end-to-end DRC/LVS validation, cannot support large multi-stage circuits such as AWGs or WDM networks, and does not integrate device- or circuit-level optical optimization. In contrast, \textsc{PICasso} is the first unified framework that couples (i) a structured NL$\rightarrow$YAML generation pipeline with PDK-aware constraints, (ii) automated placement, routing, and full physical verification, and (iii) spectrum-driven device and circuit optimization within a single closed loop. This enables \textsc{PICasso} to reliably handle all 36 tasks in PIC-Set, including 12 circuits previously unsupported by prior methods and to deliver manufacturable GDSII layouts with substantial improvements in functional Spec@\emph{k}. These distinctions establish \textsc{PICasso} as the first end-to-end autonomous PIC synthesis system rather than a netlist generator or layout assistant, closing
longstanding gaps in both robustness and physical correctness.

\begin{figure}[b]\vspace{-2em}
    \centering
    \includegraphics[width=1.01\linewidth]{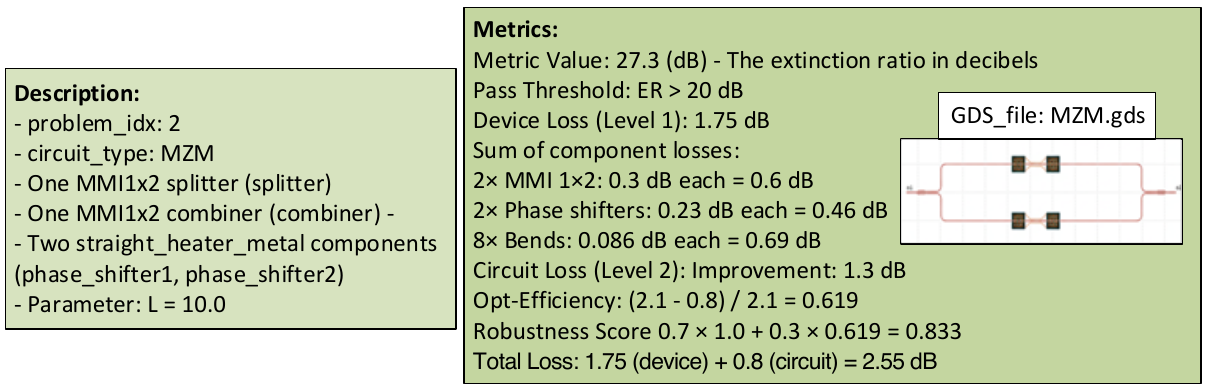} \vspace{-2em}
    \caption{Data point structure.}
    \label{datapoint_stcuture} \vspace{-1.4em}
\end{figure}

\begin{table}[t]
\centering
\caption{Comparison of \textbf{PICasso} with prior LLM-based PIC design frameworks.} 
\renewcommand{\arraystretch}{1.15}
\setlength{\tabcolsep}{3pt}
\scalebox{0.60}{
\begin{tabular}{p{2.6cm}p{2.6cm}p{2.6cm}p{5.6cm}}
\hline
\textbf{Criteria} & \textbf{PICBench~\cite{wu2025picbench}} & \textbf{PhIDO~\cite{sharma2025ai}} & \textbf{PICasso (ours)} \\
\hline
Pipeline scope & Netlist-only & Auto schematic + GDSII & NL$\rightarrow$DSL$\rightarrow$Layout$\rightarrow$Verif$\rightarrow$Opt \\
Intermediate repr. & JSON netlists & YAML$\rightarrow$gdsfactory PCells & Unified YAML for gen/route/validation \\
Dataset / tasks & 24 benchmarks & $\sim$112-component circuits & 36 parameterized components (PIC-Set) \\
Evaluation granularity & Syntax \& freq-domain & Outcomes, pass@k, runtime & Struct./func. Spec@k, optimization, robustness \\
Feedback \& repair & Error + schema checks & Pydantic + routing checks & Auto YAML check + prompt refine + opt \\
Simulation stack & SAX only & SAX + layout checks & SAX + commercial solver support \\
Layout \& routing & N/A & Basic Graphviz + DRC & Full gdsfactory PnR + DRC/LVS \\
PDK handling & SAX models & Generic Si PDK (34 PCells) & Foundry-aligned PDK w/ loss targets \\
Autonomy level & Single-agent sim & Multi-agent RAG & End-to-end auto gen + optimization \\
Typical failures & Port/element errors & Crossing/schema issues & YAML, DRC/LVS, function, optimization errors \\
Openness & Open benchmarks/scripts & Open multi-agent framework & Fully open framework + dataset + tools \\
\hline
\end{tabular}} \vspace{-2.em}
\label{tab:comparison}
\end{table}

\subsection{PIC-Set: A Curated PIC Benchmark} \label{sec:picset}

To enable standardized evaluation and fair comparison of photonic design methodologies, we introduce \textsc{PIC-Set}, a curated benchmark of representative silicon–photonic circuits constructed under a unified technology and simulation stack. The dataset emphasizes realistic design scenarios and objective diversity, covering widely used photonic primitives such as MMIs, directional couplers, phase shifters, Y-branches, and waveguide tapers that frequently appear in modulators, WDM filters, and switching fabrics.

All circuit instances are programmatically instantiated using a parameterized layout generator to ensure consistent parameter ranges, layout conventions, and PDK assumptions. Each entry includes a literature-derived target loss budget (e.g., $\sim1.3\,\mathrm{dB}$ for ring add-drop filters, $\sim2.2\,\mathrm{dB}$ for 16-channel AWGs, and $\sim0.7\,\mathrm{dB}$ for $2\times2$ thermo-optic switches), providing meaningful operating points for evaluating layout correctness and optimization performance (Fig.~\ref{datapoint_stcuture}).

To prevent information leakage across parametrically similar circuits, the dataset is partitioned at the \emph{topology} level into design, validation, and test splits. Each topology includes machine-readable YAML netlists, reference GDS/OASIS layouts, parameter bounds, structural constraints, when applicable—SPICE or S-parameter models. In addition, \textsc{PIC-Set} provides a component library with empirically grounded nominal losses for waveguides, MMIs, and directional couplers, enabling consistent evaluation across synthesis pipelines. Representative circuits and applications are summarized in Table~\ref{tab:picset-summary}.

\begin{table}[t]
  \centering
  \caption{Representative entries from \textsc{PIC-Set}. \vspace{-0.8em}}
  \label{tab:picset-summary}
  \resizebox{\columnwidth}{!}{
  \begin{tabular}{lcccc}
    \toprule
    \textbf{Circuit / Component} & \textbf{Loss Target} & \textbf{Footprint} & \textbf{Ports} & \textbf{Application} \\
    \midrule
    Ring add--drop filter & $\sim$1.3\,dB (drop) & 100--200\,µm$^2$ & 4 & WDM filtering \\
    16-ch AWG & $\sim$2.2\,dB & 0.5--1\,mm$^2$ & 32 & Multi/De-multiplexing \\
    2$\times$2 thermo-optic switch & $\sim$0.7\,dB & 200--400\,µm$^2$ & 4 & Reconfigurable routing \\
    MMI 1$\times$2 & $\sim$0.3\,dB & 20--40\,µm$^2$ & 3 & Splitting/combining \\
    Directional coupler & $\sim$0.27\,dB & 10--30\,µm$^2$ & 4 & Tapping, coherent RX \\
    Waveguide (strip) & $\sim$0.7\,dB/cm & N/A & 2 & Propagation loss \\
    MZM (push--pull) & $\sim$1.5\,dB & 0.3--0.6\,mm$^2$ & 4 & Electro-optic modulation \\
    90$^\circ$ hybrid & $\sim$1.0\,dB & $\sim$200\,µm$^2$ & 4 & Coherent detection \\
    \bottomrule
  \end{tabular}
  }\vspace{-2em}
\end{table}

\subsection{Optimization Framework}
\label{sec:optimization}
PICasso incorporates a two-tier optimization strategy to improve both device-level performance and circuit-level transmission behavior. This separation ensures that local component refinements and global interference effects are treated distinctly but cohesively.\\ \textbf{Device-Level Optimization.} At the device level, geometry and material parameters are adjusted to match target performance characteristics drawn from empirical measurements or PDK specifications. Parameters such as waveguide width and radius, MMI dimensions, and heater lengths are swept within physically realizable ranges. Analytical attenuation models and SAX-based S-parameter simulations estimate component-level losses and coupling behavior. The optimization seeks parameter vectors \(\mathbf{x}\) that minimize the deviation between simulated responses and target characteristics:
\begin{equation}
    \min_{\mathbf{x}} \;\big\|\mathbf{f}(\mathbf{x}) - \mathbf{f}_{\mathrm{target}}\big\|_2^2,
\end{equation}
where \(\mathbf{f}(\cdot)\) denotes the simulated optical response (e.g., insertion loss or coupling ratio). The result is a device configuration that adheres more closely to expected foundry behavior. \\ \textbf{Circuit-Level Optimization.} At the circuit level, the full PIC is optimized using the SAX simulator. Phase shifts and coupling coefficients across the layout are treated as tunable variables that shape multi-path interference. PICasso employs a multi-start Nelder--Mead procedure with several random initializations to avoid shallow local minima. The objective minimizes insertion loss by maximizing the squared dominant singular value \(\sigma_1^2(T)\) of the circuit transfer matrix \(T\):
\begin{equation}
    \min_{\boldsymbol{\phi}}\; -\sigma_1^2(T(\boldsymbol{\phi})),
\end{equation}
where \(\boldsymbol{\phi}\) includes all tunable phases and coupler settings. This criterion drives the circuit toward maximal throughput while preserving its functional mapping. 

\section{Evaluation Metrics}
\label{sec:metrics}
Existing PIC benchmarks rely on binary pass/fail or syntax-only correctness~\cite{wu2025picbench}, which conflate structural plausibility with physical manufacturability and cannot capture the graded improvements that a multi stage pipeline like PICasso produces. We therefore define three metrics that jointly quantify correctness, physical validity, and optimization quality across the full synthesis--verification--optimization loop. PICasso introduces a set of evaluation metrics designed to quantify correctness, physical validity, and optimization quality for automatically generated photonic circuits. These metrics capture structural validity, functional behavior, and robustness under optimization, thereby enabling  consistent comparison across PDKs, circuit topologies, and LLM backends.

\noindent\textbf{Structural Specification Satisfaction.} The first metric measures whether a generated circuit satisfies all \emph{structural} constraints imposed by the YAML DSL, the PDK, and the layout engine. A design is considered structurally valid if it satisfies the following conditions: (i) the YAML specification parses without error, (ii) the resulting component compiles through \texttt{gf.read.from\_yaml()}, (iii) automated routing completes successfully, and (iv) the final GDS passes all DRC checks under the \texttt{generic\_tech} rule deck. Layout-versus-schematic checks are included when netlist extraction is tractable. We define structural specification satisfaction at sampling depth \(k\) as \vspace{-1em}
\begin{equation}\vspace{-0.45em}
    \mathrm{Spec}\_{\mathrm{struct}}@k
    = 1 - \frac{\binom{n - c_s}{k}}{\binom{n}{k}},
\end{equation}
where \(n\) is the number of samples and \(c_s\) is the number of structurally valid designs. This formulation penalizes methods that frequently violate physical or syntactic rules, while still permitting comparison across circuits of varying complexity.

\noindent\textbf{Functional Specification Satisfaction.} For designs that pass structural checks, PICasso evaluates whether the resulting circuit satisfies the intended photonic functionality using SAX-based frequency-domain simulation. Functional criteria include power transfer, insertion loss trends, wavelength selectivity, and, when applicable, switching or splitting ratios. A design passes functional validation if its simulated behavior lies within topology-specific tolerances derived from \textsc{PIC-Set}. We define functional specification satisfaction as \vspace{-1em}
\begin{equation}\vspace{-0.4em}
    \mathrm{Spec}\_{\mathrm{full}}@k
    = 1 - \frac{\binom{n - c_f}{k}}{\binom{n}{k}},
\end{equation} 
where \(c_f\) counts the number of designs that satisfy both structural and functional requirements. This metric distinguishes PICasso from baseline LLM outputs, which seldom meet functional objectives despite occasional syntactic correctness. 

\noindent\textbf{Optimization Efficiency.} Optimization effectiveness is evaluated by measuring the normalized reduction in insertion loss between the initial layout and the optimized configuration. Let \(IL_{\mathrm{before}}\) and \(IL_{\mathrm{after}}\) denote the insertion loss before and after optimization. We define
$    \eta_{\mathrm{opt}}
    = \frac{IL_{\mathrm{before}} - IL_{\mathrm{after}}}
           {IL_{\mathrm{before}} + \varepsilon}$, where \(\varepsilon\) is where $\varepsilon = 10^{-6}$ prevents division by near-zero denominators. used to avoid division by arbitrarily small denominators. The metric is clipped to the interval \([0,1]\) and averaged across all circuits that pass structural validation. This reflects the gain attributable to PICasso’s device- and circuit-level tuning.

\begin{table*}[t] \vspace{-1em} 
\centering
\caption{Aggregate Spec@k performance across all 36 PIC-Set tasks.} 
\label{tab:picasso_all36}
\renewcommand{\arraystretch}{1.05}
\setlength{\tabcolsep}{3pt}
\scalebox{0.9}{
\begin{tabular}{c*{12}{c}}
\toprule
\multirow{3}{*}{\textbf{Model / Setting}}
& \multicolumn{4}{c}{\textbf{Complexity 1}}
& \multicolumn{4}{c}{\textbf{Complexity 2}}
& \multicolumn{4}{c}{\textbf{Complexity 3}} \\
\cmidrule(lr){2-5}\cmidrule(lr){6-9}\cmidrule(lr){10-13}
& \multicolumn{2}{c}{\textbf{Structural}} & \multicolumn{2}{c}{\textbf{Full Spec}}
& \multicolumn{2}{c}{\textbf{Structural}} & \multicolumn{2}{c}{\textbf{Full Spec}}
& \multicolumn{2}{c}{\textbf{Structural}} & \multicolumn{2}{c}{\textbf{Full Spec}} \\
\cmidrule(lr){2-3}\cmidrule(lr){4-5}
\cmidrule(lr){6-7}\cmidrule(lr){8-9}
\cmidrule(lr){10-11}\cmidrule(lr){12-13}
& \textbf{Spec@1} & \textbf{Spec@3} & \textbf{Spec@1} & \textbf{Spec@3}
& \textbf{Spec@1} & \textbf{Spec@3} & \textbf{Spec@1} & \textbf{Spec@3}
& \textbf{Spec@1} & \textbf{Spec@3} & \textbf{Spec@1} & \textbf{Spec@3} \\
\midrule
GPT-4o
& \cellcolor{red!25}46.7 & \cellcolor{red!25}62.2 & 0.0 & 0.0
& \cellcolor{red!25}31.7 & \cellcolor{red!25}33.3 & 0.0 & 0.0
& 0.0 & 0.0 & 0.0 & 0.0 \\
\quad PICasso w/ GPT-4o
& \cellcolor{green!25}97.8 & \cellcolor{green!25}100.0 & \cellcolor{green!25}88.9 & \cellcolor{green!25}88.9
& \cellcolor{green!25}95.0 & \cellcolor{green!25}99.2 & \cellcolor{yellow!25}80.0 & \cellcolor{green!25}90.8
& \cellcolor{yellow!25}48.0 & \cellcolor{yellow!25}63.3 & \cellcolor{red!25}8.0 & \cellcolor{red!25}14.7 \\
Claude Sonnet 4.5
& \cellcolor{red!25}22.2 & \cellcolor{red!25}28.9 & 0.0 & 0.0
& \cellcolor{red!25}33.3 & \cellcolor{red!25}52.5 & 0.0 & 0.0
& \cellcolor{red!25}8.0 & \cellcolor{red!25}10.7 & 0.0 & 0.0 \\
\quad PICasso w/ Claude Sonnet 4.5
& \cellcolor{green!25}100.0 & \cellcolor{green!25}100.0 & \cellcolor{green!25}97.8 & \cellcolor{green!25}100.0
& \cellcolor{green!25}100.0 & \cellcolor{green!25}100.0 & \cellcolor{green!25}98.3 & \cellcolor{green!25}100.0
& \cellcolor{yellow!25}84.0 & \cellcolor{yellow!25}90.7 & \cellcolor{yellow!25}25.3 & \cellcolor{yellow!25}52.0 \\
Llama 3.1 70B
& \cellcolor{red!25}33.3 & \cellcolor{red!25}48.9 & 0.0 & 0.0
& 0.0 & 0.0 & 0.0 & 0.0
& \cellcolor{red!25}2.7 & \cellcolor{red!25}6.0 & 0.0 & 0.0 \\
\quad PICasso w/ Llama 3.1 70B
& \cellcolor{green!25}80.0 & \cellcolor{green!25}94.4 & \cellcolor{yellow!25}68.9 & \cellcolor{yellow!25}80.0
& \cellcolor{green!25}71.7 & \cellcolor{green!25}96.7 & \cellcolor{yellow!25}33.3 & \cellcolor{yellow!25}50.0
& \cellcolor{green!25}77.3 & \cellcolor{green!25}92.7 & 0.0 & 0.0 \\
\bottomrule
\end{tabular}\vspace{-1em}
}
\end{table*}

\section{Experiment results}
\label{sec:experiments}

\noindent\textbf{Setup.} Evaluation follows a two-phase protocol. Phase~1 (\textit{vanilla}) runs each LLM with no structural priors, routing constraints, or optimization, measuring inherent generation capability. Phase~2 (\textit{framework}) activates the full PICasso pipeline: schema checking, DRC/LVS enforcement, SAX-based functional evaluation, and two-stage optimization. For each of the 36 PIC-Set tasks, five samples are drawn per phase(180 samples per model per phase). Complexity tiers follow the three-level taxonomy defined in Sec.~\ref{sec:picset}: C1 (single-component), C2 (2--8 components), C3 ($>$8 components).

For fair comparison to prior literature, the 24 PICBench compatible tasks form a secondary benchmark subset. Because existing frameworks differ substantially in their input/output representations, physical validation scope, and simulation stacks, direct numerical comparison is inappropriate. Instead, a capability-level comparison is provided in Table \ref{tab:frameworks}, and focuses on design-space coverage, representation, automation, and runtime.

\noindent$\textbf{Results and Analysis.}$ Results are reported using $\text{Spec@}k$ metrics, which reflect end-to-end specification satisfaction.The metric, $\text{Spec@1\_structural}$ reports the probability that the top-1 sample satisfies structural constraints, while $\text{Spec@3\_structural}$ measures correctness across the best of 3. $\text{Spec@1\_full}$ and $\text{Spec@3\_full}$ additionally require functional correctness. This is stricter than prior work, as it accounts for both physical-level (DRC/LVS) and reflects final post-optimization behavior.

\begin{figure}[t]\vspace{-1em}
    \centering
    \includegraphics[width=1.01\linewidth]{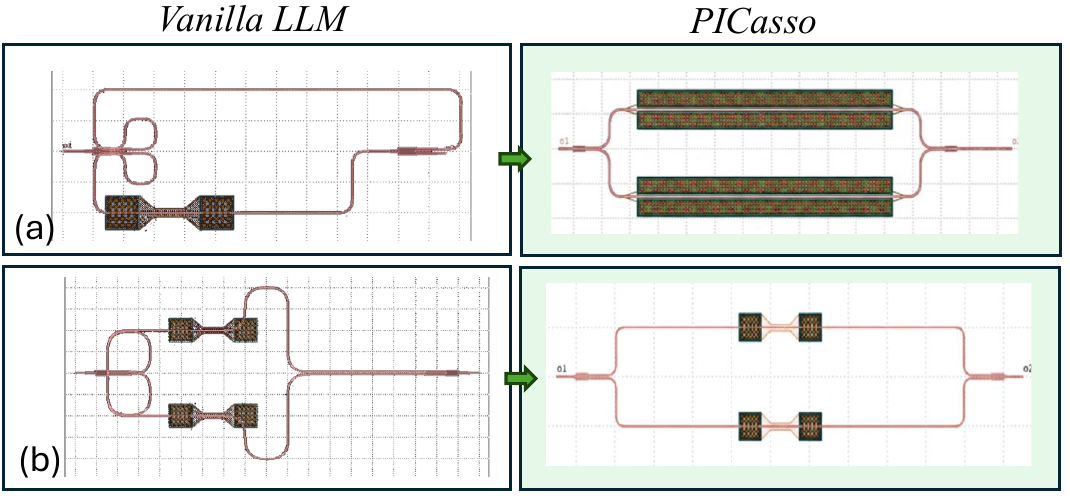} \vspace{-2em}
    \caption{Comparison between Vanilla LLM and PICasso on (a) MZI (b) MZM layout generation.}
    \label{layput} \vspace{-1em}
\end{figure}

Table~\ref{tab:picasso_all36} reports aggregate results across all 36 PIC-Set tasks for three state-of-the-art LLMs: GPT-4o, Claude Sonnet 4.5, and Llama 3.1 70B. Results are reported for both vanilla generation and the full PICasso pipeline.

Across all models and complexity tiers, PICasso substantially improves satisfaction with structural and functional specifications. For example, GPT-4o achieves structural $\mathrm{Spec}@3$ of $99.2\%$ on medium-complexity circuits under PICasso, compared to $33.3\%$ under vanilla generation. Claude Sonnet 4.5 shows the strongest performance, achieving $100\%$ structural $\mathrm{Spec}@3$ and $100\%$ functional $\mathrm{Spec}@3$ on Complexity-2 tasks. Even for large circuits in Complexity-3, PICasso improves structural $\mathrm{Spec}@3$ to $90.7\%$ and functional $\mathrm{Spec}@3$ to $52.0\%$, whereas vanilla LLM generation fails to produce functionally correct designs.

Overall, the results indicate that PICasso transforms LLMs from unreliable circuit generators into practical PIC design agents capable of producing manufacturable layouts that satisfy both physical and functional constraints.

A cross-framework comparison (Table \ref{tab:frameworks}) highlights qualitative differences among PICBench, PhIDO, and PICasso. PICBench supports 24 netlist-only tasks and evaluates frequency-domain behavior but lacks layout or DRC/LVS capabilities. PhIDO extends this to multi-agent NL-to-layout generation but supports fewer physical checks and cannot handle AWG, WDM, or large multi-stage circuits. PICasso supports 36 tasks, including 12 circuits that were previously unsupported, and integrates full physical validation, spectral evaluation, and optimization. Despite deeper validation, the runtime remains competitive with prior multi-agent frameworks due to targeted repair and parallelized checking.
PICasso therefore establishes the first end-to-end autonomous PIC synthesis framework capable of natural-language input, structured multilayer validation, and functional optimization, with demonstrated improvements across diverse LLM families. The results indicate that architectural priors, validation-loop feedback, and physical-level constraints are crucial for PIC code generation and cannot be achieved solely through prompting.

\begin{table}[t]
\centering
\caption{Experiment runtime comparison.}
\label{tab:frameworks}
\vspace{-0.75em}
\renewcommand{\arraystretch}{1.25}
\setlength{\tabcolsep}{4pt}
\scalebox{0.78}{
    \begin{threeparttable}
    \begin{tabular}{l p{2.6cm} p{4.0cm} p{2.0cm} p{2.2cm}}
    \hline
    \textbf{Method} & \textbf{Task Coverage} & \textbf{Representation / Pipeline Depth} & \textbf{Runtime (min)} \\
    \hline
    PICBench~\cite{wu2025picbench} & 24 tasks & JSON Netlists $\rightarrow$ SAX simulation & 1--3 \\
    PhIDO~\cite{sharma2025ai} & $\sim$102 tasks & NL $\rightarrow$ YAML $\rightarrow$ Layout & 3--6 \\
    \textbf{PICasso (ours)} & \textbf{36 tasks} & \textbf{Unified YAML $\rightarrow$ GDS $\rightarrow$ Optimization} & \textbf{4--8} \\
    \textit{Human Designer\tnote{1}} & PIC-Set & Drag-and-place PCells + routing & \textit{25--60} \\
    \hline
    \end{tabular}

    \begin{tablenotes}[flushleft]
    \footnotesize 
    \item[1] ``Human Designer'' refers to a trained PIC engineer implementing the same tasks in a commercial photonic design GUI (e.g., IPKISS, Nazca, KLayout+GDSFactory PCells). PICasso achieves full-stack verification and supports higher-complexity circuits while eliminating virtually all manual effort.
    \end{tablenotes}
    \end{threeparttable}
} 

\vspace{-1.5em}
\end{table}

\subsection{Optimization analysis}
\begin{figure}[t]
\centering
\includegraphics[width=0.95\linewidth]{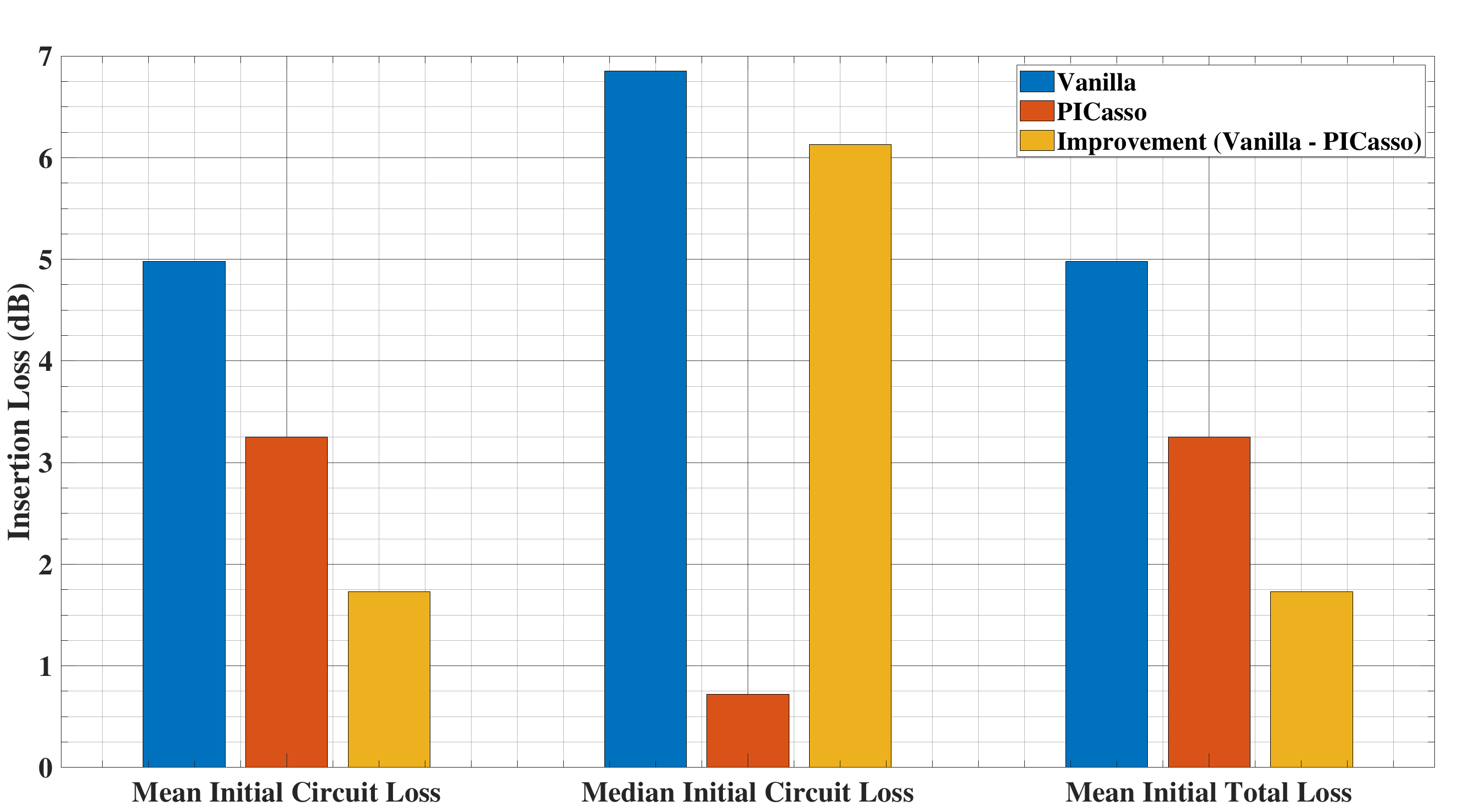}\vspace{-0.5em}
\caption{Initial Loss Comparison: Vanilla vs PICasso ($N\_V$ = 48, $N\_P$ = 158).} \vspace{-2em}
\label{fig:initial-loss-comparison}
\end{figure}

Fig.~\ref{fig:initial-loss-comparison} reports the initial optical insertion loss achieved by vanilla LLM-based circuit generation and by the proposed \textsc{PICasso} framework, prior to any explicit optimization. The figure is constructed from pooled summary statistics computed over all circuits that successfully build and pass structural validation across five LLM back-ends (claude-sonnet-4.5, deepseek-r1-api, gpt-4o, llama-3.1-70b, and qwen-2.5-32b).

The underlying data originate from the optimization, which record the pre- and post-optimization losses for each generated circuit. To ensure a fair comparison, we apply identical filtering criteria to the vanilla and \textsc{PICasso} phases:
(i) only samples with \texttt{structural\_pass = true} are retained, meaning the circuit can be instantiated from the generated YAML and passes DRC checks;
(ii) we require that a YAML or GDS file is available so that loss can be evaluated;
(iii) we restrict to models that have at least one vanilla and one \textsc{PICasso} sample after filtering. After these steps, we obtain $48$ valid vanilla circuits and $158$ valid \textsc{PICasso} circuits (206 circuits in total).
Let $\mathcal{V}$ and $\mathcal{P}$ denote the sets of valid vanilla and \textsc{PICasso} samples, respectively, and let $L^{\text{circ}}_{\mathrm{before}}$ and $L^{\text{tot}}_{\mathrm{before}}$ denote the initial circuit and total insertion loss in decibels. We pool all samples across models and compute population-level statistics for each phase:
the mean and median initial circuit loss for vanilla and \textsc{PICasso}, as well as the mean initial total loss. The improvement shown in Fig.~\ref{fig:initial-loss-comparison} is defined as the difference between phases,
$\Delta = \text{Vanilla} - \text{\textsc{PICasso}}$,
so $\Delta > 0$ indicates that \textsc{PICasso} yields lower loss.

Numerically, the pooled mean initial circuit loss decreases from $4.98$~dB for vanilla to $3.25$~dB for \textsc{PICasso}, corresponding to an average improvement of $1.74$~dB. The median initial circuit loss drops from $6.85$~dB to $0.72$~dB, a $6.13$~dB reduction that reflects the large number of \textsc{PICasso}-generated circuits with near-zero insertion loss. Because device loss is $0$~dB for both phases in this analysis, the mean initial total loss matches the circuit-loss statistics and also improves by $1.74$~dB. Taken together, Fig.~\ref{fig:initial-loss-comparison} shows that, among circuits that successfully build and pass structural checks, the \textsc{PICasso} framework consistently produces higher-quality initial photonic circuits with substantially lower optical insertion loss across all five LLM models.

\noindent\textbf{Execution time comparison.} To highlight the additional capabilities of our framework and its minimal human-effort requirements, we compare execution times across prior approaches. As shown in Table~\ref{tab:frameworks}, PICasso achieves full-stack verification and supports higher-complexity circuits while maintaining competitive runtimes. Despite performing deeper pipeline steps than netlist- or schematic-level baselines, PICasso completes tasks in 4–8 minutes, substantially faster than manual GUI-based design workflows, which require 25–60 minutes. 

\subsection{Demonstration of PICasso}\vspace{-0.25em}
\label{subsec:pipeline-contribution}
 
\noindent Table~\ref{tab:ablation} isolates the contribution of each PICasso stage on 12 representative PIC-Set tasks (Claude Sonnet~4.5, $n{=}5$, Spec@5).
Raw and knowledge-injected generation (Baseline, V1--V2) reaches at most 25--42\% structural correctness and zero functional Spec@5, as LLM output routinely fails port-consistency and routing checks before reaching the layout engine.
Pilot validation (V3) alone saturates structural correctness at 100\% across all tiers; physical and functional verification (V4) then raises full-specification Spec@5 from 0\% to 91.7\%, with device- and circuit-level optimisation (V5, full PICasso) yielding 83.3\%---consistent with the aggregate gains in Table~\ref{tab:picasso_all36}.
The 64-QAM Modulator (C3) and 90$^\circ$ Optical Hybrid (C2) achieve 100\% structural correctness under V5 but zero functional Spec@5, as their phase-matching constraints exceed the convergence radius of the Nelder--Mead optimizer; improving optimizer scope for high-dimensional photonic circuits remains an open direction.

\begin{table}[t]
\centering
\caption{PICasso pipeline contribution on 12 PIC-Set tasks
  (Claude Sonnet~4.5, $n{=}5$, Spec@5).}
\label{tab:ablation}
\renewcommand{\arraystretch}{1.12}
\setlength{\tabcolsep}{4pt}
\scalebox{0.90}{%
\begin{tabular}{>{\raggedright\arraybackslash}p{3.9cm} c r r r r}
\toprule
\textbf{Problem} & \textbf{Tier}
  & \cellcolor{hBase}\textbf{Base S}$^\dagger$
  & \cellcolor{hV3}\textbf{V3 S}
  & \cellcolor{hV4}\textbf{V4 F}
  & \cellcolor{hV5}\textbf{V5 F} \\
\midrule
MZI (Mach-Zehnder Interf.)  & \TC{1} & \Szero  & \Sfull & \Sfull & \Sfull \\
MZM (Mach-Zehnder Mod.)     & \TC{1} & \Szero  & \Sfull & \Sfull & \Sfull \\
2$\times$2 Optical Switch   & \TC{1} & \Szero  & \Sfull & \Sfull & \Sfull \\
MMI 1$\times$2 Splitter     & \TC{1} & \Sfull  & \Sfull & \Sfull & \Sfull \\
\midrule
QPSK Modulator              & \TC{2} & \Szero  & \Sfull & \Sfull & \Sfull \\
90$^\circ$ Optical Hybrid   & \TC{2} & \Sfull  & \Sfull & \Sfull & \Szero \\
Ring Add-Drop Filter        & \TC{2} & \Sfull  & \Sfull & \Sfull & \Sfull \\
Tunable 1$\times$4 Switch   & \TC{2} & \Sfull  & \Sfull & \Szero & \Sfull \\
\midrule
64-QAM Modulator            & \TC{3} & \Sfull  & \Sfull & \Sfull & \Szero \\
4$\times$4 Crossbar Switch  & \TC{3} & \Sfull  & \Sfull & \Sfull & \Sfull \\
Clements 4$\times$4 Mesh    & \TC{3} & \Sfull  & \Sfull & \Sfull & \Sfull \\
4-ch.\ WDM Cascaded MZI     & \TC{3} & \Szero  & \Sfull & \Sfull & \Sfull \\
\midrule
\multicolumn{2}{l}{\textit{Avg C1}}
  & \Szero       & \Sfull & \Sfull       & \Sfull      \\
\multicolumn{2}{l}{\textit{Avg C2}}
  & \Spart{50.0} & \Sfull & \Spart{75.0} & \Spart{75.0}\\
\multicolumn{2}{l}{\textit{Avg C3}}
  & \Spart{75.0} & \Sfull & \Sfull       & \Spart{75.0}\\
\midrule
\multicolumn{2}{l}{\textbf{Overall}}
  & \cellcolor{cPart}\textbf{25--42}
  & \cellcolor{cFull}\textbf{100.0}
  & \cellcolor{cPart}\textbf{91.7}
  & \cellcolor{cPart}\textbf{83.3} \\
\bottomrule
\multicolumn{6}{l}{%
  \scriptsize
  $^\dagger$Base: best S over V1 (vanilla) and V2 (+YAML/knowledge injection),
  range 25--42\%.}\\
\multicolumn{6}{l}{%
  \scriptsize
  V3: +pilot validation;\; V4: +DRC/LVS/SAX;\; V5: full PICasso.}\\
\multicolumn{6}{l}{%
  \scriptsize
  S$=100\%$ from V3 onward (col.\ omitted for V4--V5);\;
  F undefined for V1--V3 by design.}\\
\multicolumn{6}{l}{%
  \scriptsize
  \colorbox{cZero}{~~}0\%\quad\colorbox{cPart}{~~}partial\quad
  \colorbox{cFull}{~~}100\%.}\\
\end{tabular}}\vspace{-2em}
\end{table}

\section{Conclusion}
This work introduced \textsc{PICasso}, an end-to-end LLM-driven framework for autonomous PIC synthesis, verification, and optimization, together with \textsc{PIC-Set}, a curated benchmark of 36 parameterized silicon-photonic design tasks. By coupling a PDK-aware NL$\rightarrow$YAML$\rightarrow$GDS pipeline with DRC/LVS checks, SAX-based simulation, and a two-stage optimization loop, PICasso reliably generates manufacturable layouts from natural-language intent. The proposed metrics, structural and functional $\mathrm{Spec}@k$ and optimization efficiency, enable rigorous, reproducible evaluation across models, topologies, and complexity tiers. Experiments across GPT-4o, Claude Sonnet~4.5, and Llama~3.1~70B demonstrate that PICasso raises full-spec $\mathrm{Spec}@3$ from 0\% to up to 52\% on high-complexity circuits, achieves 100\% structural and functional $\mathrm{Spec}@3$ on Complexity~1 and~2 under Claude Sonnet~4.5, and reduces mean insertion loss from 4.98\,dB to 3.25\,dB (1.74\,dB improvement) all within 4--8 minutes per task versus 25--60 minutes for manual GUI-based workflows.Remaining limitations include functional failures on phase-sensitive high-complexity circuits (e.g., 64-QAM, 90$^\circ$ hybrid) where Nelder--Mead convergence is insufficient; extending the optimizer to gradient-based or learned surrogate methods is the primary open direction.
\vspace{-1.05em}

\bibliographystyle{IEEEtran}
\bibliography{Reference}\vspace{-2em}

@inproceedings{10323953,
  author       = {Fu, Y. and others},
  title        = {Gpt4aigchip: Towards next-generation ai accelerator design automation via large language models},
  booktitle    = {2023 IEEE/ACM International Conference on Computer Aided Design (ICCAD)},
  year         = {2023},
  pages        = {1--9},
  organization = {IEEE}
}

@article{vungarala2024spicepilot,
  author       = {Vungarala, D. and others},
  title        = {Spicepilot: Navigating spice code generation and simulation with ai guidance},
  journal      = {arXiv preprint arXiv:2410.20553},
  year         = {2024}
}

@article{lai2024analogcoder,
  author       = {Lai, Y. and others},
  title        = {Analogcoder: Analog circuit design via training-free code generation},
  journal      = {arXiv preprint arXiv:2405.14918},
  year         = {2024}
}

@article{thakur2023verigen,
  author       = {Thakur, S. and others},
  title        = {Verigen: A large language model for verilog code generation},
  journal      = {ACM Transactions on Design Automation of Electronic Systems},
  year         = {2024},
  volume       = {29},
  number       = {3},
  pages        = {1--31}
}

@article{he2023chateda,
  author       = {Wu, H. and others},
  title        = {Chateda: A large language model powered autonomous agent for eda},
  journal      = {IEEE Transactions on Computer-Aided Design of Integrated Circuits and Systems},
  year         = {2024}
}

@article{hu2024uvllm,
  author       = {Hu, Y. and others},
  title        = {Uvllm: An automated universal rtl verification framework using llms},
  journal      = {arXiv preprint arXiv:2411.16238},
  year         = {2024}
}

@article{chang2023chipgpt,
  author       = {Chang, K. and others},
  title        = {Chipgpt: How far are we from natural language hardware design},
  journal      = {arXiv preprint arXiv:2305.14019},
  year         = {2023}
}

@inproceedings{zhang2024mgverilog,
  author       = {Zhang, Y. and others},
  title        = {{MG-Verilog: Multi-grained Dataset Towards Enhanced LLM-assisted Verilog Generation}},
  booktitle    = {2024 IEEE LLM Aided Design Workshop (LAD)},
  year         = {2024}
}

@article{nadimi2024multi,
  author       = {Nadimi, B. and Zheng, H.},
  title        = {A multi-expert large language model architecture for verilog code generation},
  journal      = {arXiv preprint arXiv:2404.08029},
  year         = {2024}
}

@article{li2025deepcircuitx,
  author       = {Li, Z. and others},
  title        = {Deepcircuitx: A comprehensive repository-level dataset for rtl code understanding, generation, and ppa analysis},
  journal      = {arXiv preprint arXiv:2502.18297},
  year         = {2025}
}

@inproceedings{lu2023rtllm,
  author       = {Lu, Y. and others},
  title        = {Rtllm: An open-source benchmark for design rtl generation with large language model},
  booktitle    = {2024 29th Asia and South Pacific Design Automation Conference (ASP-DAC)},
  year         = {2024},
  pages        = {722--727},
  organization = {IEEE}
}

@inproceedings{zhang2024llmcompass,
  author       = {Zhang, H. and others},
  title        = {Llmcompass: Enabling efficient hardware design for large language model inference},
  booktitle    = {2024 ACM/IEEE 51st Annual International Symposium on Computer Architecture (ISCA)},
  year         = {2024},
  pages        = {1080--1096},
  organization = {IEEE}
}

@article{dai2022can,
  author       = {Dai, D. and others},
  title        = {Why can gpt learn in-context? language models implicitly perform gradient descent as meta-optimizers},
  journal      = {arXiv preprint arXiv:2212.10559},
  year         = {2022}
}

@online{bhandari2025masalachailargescalespicenetlist,
  author       = {Bhandari, J. and others},
  title        = {Masala-chai: A large-scale spice netlist dataset for analog circuits by harnessing ai},
  year         = {2025},
  url          = {https://arxiv.org/abs/2411.14299}
}

@article{fang2024assertllm,
  title={AssertLLM: Generating and Evaluating Hardware Verification Assertions from Design Specifications via Multi-LLMs},
  author={Fang, Wenji and Li, Mengming and Li, Min and Yan, Zhiyuan and Liu, Shang and Zhang, Hongce and Xie, Zhiyao},
  journal={arXiv preprint arXiv:2402.00386},
  year={2024}
}

@article{thakur2023autochip,
  title={Autochip: Automating hdl generation using llm feedback},
  author={Thakur, Shailja and Blocklove, Jason and Pearce, Hammond and Tan, Benjamin and Garg, Siddharth and Karri, Ramesh},
  journal={arXiv preprint arXiv:2311.04887},
  year={2023}
}

@inproceedings{blocklove2023chip,
  title={Chip-chat: Challenges and opportunities in conversational hardware design},
  author={Blocklove, Jason and Garg, Siddharth and Karri, Ramesh and Pearce, Hammond},
  booktitle={2023 ACM/IEEE 5th Workshop on Machine Learning for CAD (MLCAD)},
  pages={1--6},
  year={2023},
  organization={IEEE}
}

@misc{ansyslumerical,
  author       = {ANSYS Inc.},
  title        = {Lumerical Photonic Design Suite},
  howpublished = {\url{https://www.lumerical.com}},
  year         = {2025},
  note         = {Version [specify version used, e.g., 2023 R2]}
}

@misc{sax,
  author       = {Alexander Lukaszewicz and Wouter Bogaerts},
  title        = {SAX: Scalable Circuit Modeling for Photonic Design},
  year         = {2023},
  publisher    = {Zenodo},
  version      = {0.13.2},
  doi          = {10.5281/zenodo.7149361},
  url          = {https://doi.org/10.5281/zenodo.7149361}
}

@misc{ansys,
  author       = {ANSYS Inc.},
  title        = {ANSYS Multiphysics Simulation Software},
  howpublished = {ANSYS Inc., Canonsburg, PA, USA. Available at: \url{https://www.ansys.com}},
  year         = {2025},
  note         = {Version 2025 R1}
}

@article{vungarala2025limca,
  title={LIMCA: LLM for Automating Analog In-Memory Computing Architecture Design Exploration},
  author={Vungarala, Deepak and Amin, Md Hasibul and Mercati, Pietro and Ghosh, Arnob and Roohi, Arman and Zand, Ramtin and Angizi, Shaahin},
  journal={arXiv preprint arXiv:2503.13301},
  year={2025}
}

@unpublished{li:hal-04175312,
  TITLE = {{From English to PCSEL: LLM helps design and optimize photonic crystal surface emitting lasers}},
  AUTHOR = {Li, Renjie and Zhang, Ceyao and Mao, Sixuan and Huang, Hai and Zhong, Mou and Cui, Yiou and Zhou, Xiyuan and Yin, Feng and Theodoridis, Sergios and Zhang, Zhaoyu},
  URL = {https://hal.science/hal-04175312},
  NOTE = {14 pages, 9 graphics},
  HAL_LOCAL_REFERENCE = {to be updated},
  YEAR = {2023},
  MONTH = Aug,
  DOI = {10.48550/arXiv.2104.12145},
  HAL_ID = {hal-04175312},
  HAL_VERSION = {v2},
}

@inproceedings{liu2024towards,
  title={Towards large-language model assisted layout of silicon photonic integrated circuits},
  author={Liu, Jason and Sharma, Ankita and Doumbia, Cheick and Poon, Joyce KS},
  booktitle={European Conference on Integrated Optics},
  pages={441--447},
  year={2024},
  organization={Springer}
}

@inproceedings{wu2025picbench,
  title={PICBench: Benchmarking LLMs for Photonic Integrated Circuits Design},
  author={Wu, Yuchao and Yu, Xiaofei and Chen, Hao and Luo, Yang and Tong, Yeyu and Ma, Yuzhe},
  booktitle={2025 Design, Automation \& Test in Europe Conference (DATE)},
  pages={1--6},
  year={2025},
  organization={IEEE}
}

@misc{marketsandmarkets2025,
  author = {{MarketsandMarkets}},
  title = {{Silicon Photonics Market by Product, Components - Global Forecast to 2030}},
  howpublished = {\url{https://www.marketsandmarkets.com/Market-Reports/silicon-photonics-116.html}},
  year = {2025},
  note = {[Accessed: 20-October-2025]}
}

@article{shen2024recent,
  title={Recent progress on inverse design for integrated photonic devices: methodology and applications},
  author={Shen, Ruoyu and Hong, Bingzhou and Ren, Xiuyan and Yang, Fenghe and Chu, Wei and Cai, Haiwen and Huang, Weiping},
  journal={Journal of Nanophotonics},
  volume={18},
  number={1},
  pages={010901--010901},
  year={2024},
  publisher={Society of Photo-Optical Instrumentation Engineers}
}

@article{baek2025recent,
  title={Recent advances in the inverse design of silicon photonic devices and related platforms using deep generative models},
  author={Baek, Sun Jae and Lee, Minhyeok},
  journal={PeerJ Computer Science},
  volume={11},
  pages={e2895},
  year={2025},
  publisher={PeerJ Inc.}
}

@inproceedings{zhou2025toward,
  title={Toward intelligent electronic-photonic design automation for large-scale photonic integrated circuits: from device inverse design to physical layout generation},
  author={Zhou, Hongjian and Ma, Pingchuan and Gu, Jiaqi},
  booktitle={Optical Design Automation},
  volume={13601},
  pages={69--78},
  year={2025},
  organization={SPIE}
}

@misc{gdsfactory,
  author       = {Matres, Joaquin and GDSFactory Contributors},
  title        = {{GDSFactory: Python-Based Photonic and Analog Chip Design Automation}},
  howpublished = {\url{https://gdsfactory.github.io/gdsfactory/}},
  year         = {2025},
  note         = {Version 9.x. Accessed: 2025-11-11}
}

@article{sharma2025ai,
  title={AI Agents for Photonic Integrated Circuit Design Automation},
  author={Sharma, Ankita and Fu, YuQi and Ansari, Vahid and Iyer, Rishabh and Kuang, Fiona and Mistry, Kashish and Aishy, Raisa Islam and Ahmad, Sara and Matres, Joaquin and Englund, Dirk R and others},
  journal={arXiv preprint arXiv:2508.14123},
  year={2025}
}

@ARTICLE{layoutcopilot,
  author={Liu, Bingyang and Zhang, Haoyi and Gao, Xiaohan and Kong, Zichen and Tang, Xiyuan and Lin, Yibo and Wang, Runsheng and Huang, Ru},
  journal={IEEE Transactions on Computer-Aided Design of Integrated Circuits and Systems}, 
  title={LayoutCopilot: An LLM-Powered Multiagent Collaborative Framework for Interactive Analog Layout Design}, 
  year={2025},
  volume={44},
  number={8},
  pages={3126-3139},
  doi={10.1109/TCAD.2025.3529805}}

@INPROCEEDINGS{interactive1,
  author={Gao, Xiaohan and Liu, Mingjie and Pan, David Z. and Lin, Yibo},
  booktitle={2021 58th ACM/IEEE Design Automation Conference (DAC)}, 
  title={Interactive Analog Layout Editing with Instant Placement Legalization}, 
  year={2021},
  volume={},
  number={},
  pages={1249-1254},
  doi={10.1109/DAC18074.2021.9586234}}

@ARTICLE{interactive2,
  author={Gao, Xiaohan and Zhang, Haoyi and Liu, Mingjie and Shen, Linxiao and Pan, David Z. and Lin, Yibo and Wang, Runsheng and Huang, Ru},
  journal={IEEE Transactions on Computer-Aided Design of Integrated Circuits and Systems}, 
  title={Interactive Analog Layout Editing With Instant Placement and Routing Legalization}, 
  year={2023},
  volume={42},
  number={3},
  pages={698-711},
  doi={10.1109/TCAD.2022.3190234}}

@INPROCEEDINGS{11106005,
  author={Vungarala, Deepak and Elbtity, Mohammed Essa and Pandit, Kartik and Syed, Sumiya and Alam, Sakila and Ghosh, Arnob and Zand, Ramtin and Angizi, Shaahin},
  booktitle={2025 IEEE International Conference on LLM-Aided Design (ICLAD)}, 
  title={TPU-Gen: LLM-Driven Custom Tensor Processing Unit Generator}, 
  year={2025},
  volume={},
  number={},
  pages={1-8},
  doi={10.1109/ICLAD65226.2025.00010}}

@article{khan2019photonic,
  title={Photonic integrated circuit design in a foundry+ fabless ecosystem},
  author={Khan, Muhammad Umar and Xing, Yufei and Ye, Yinghao and Bogaerts, Wim},
  journal={IEEE Journal of Selected Topics in Quantum Electronics},
  volume={25},
  number={5},
  pages={1--14},
  year={2019},
  publisher={IEEE}
}

@article{maclellan2024inverse,
  title={Inverse design of photonic systems},
  author={MacLellan, Benjamin and Roztocki, Piotr and Belleville, Julie and Romero Cort{\'e}s, Luis and Ruscitti, Kaleb and Fischer, Bennet and Aza{\~n}a, Jos{\'e} and Morandotti, Roberto},
  journal={Laser \& Photonics Reviews},
  volume={18},
  number={5},
  pages={2300500},
  year={2024},
  publisher={Wiley Online Library}
}

@article{li2023deep,
  title={Deep reinforcement learning empowers automated inverse design and optimization of photonic crystals for nanoscale laser cavities},
  author={Li, Renjie and Zhang, Ceyao and Xie, Wentao and Gong, Yuanhao and Ding, Feilong and Dai, Hui and Chen, Zihan and Yin, Feng and Zhang, Zhaoyu},
  journal={Nanophotonics},
  volume={12},
  number={2},
  pages={319--334},
  year={2023},
  publisher={De Gruyter}
}

\end{document}